\documentclass[letterpaper, 10pt, conference]{IEEEtran}
\usepackage{times}
\usepackage{xr}
\usepackage{amsmath,amssymb,mathrsfs}
\usepackage[]{algorithm2e}
\usepackage{graphicx}
\usepackage[table]{xcolor}
\usepackage{wrapfig}
\usepackage{textcomp}
\usepackage{booktabs}
\usepackage{multirow}
\usepackage{textcomp}
\usepackage{textcomp} 

\definecolor{control}{RGB}{203, 65, 107}
\definecolor{shape}{RGB}{61, 153, 115}

\usepackage[numbers]{natbib}
\usepackage{multicol}
\usepackage[bookmarks=true, hidelinks, colorlinks = true, linkcolor = black, urlcolor  = blue, citecolor = black, linkcolor=black]{hyperref}

\begin{document}

\title{\vspace{5pt}Scalable sim-to-real transfer of soft robot designs}

\author{Author Names Omitted for Anonymous Review. Paper-ID [add your ID here]}


\author{
\IEEEauthorblockN{Sam Kriegman$^1$,\quad
Amir Mohammadi Nasab$^2$,\quad
Dylan Shah$^2$,\quad
Hannah Steele$^2$,\quad
Gabrielle Branin$^2$,\\[2pt]
Michael Levin$^3$,\quad
Josh Bongard$^1$,\quad
Rebecca Kramer-Bottiglio$^2$}\vspace{6pt}
\IEEEauthorblockA{$^1$University of Vermont,  $^2$Yale University,  $^3$Tufts University}
}

\teaser{
\centering
\vspace{-2pt}
    \includegraphics[width=\linewidth]{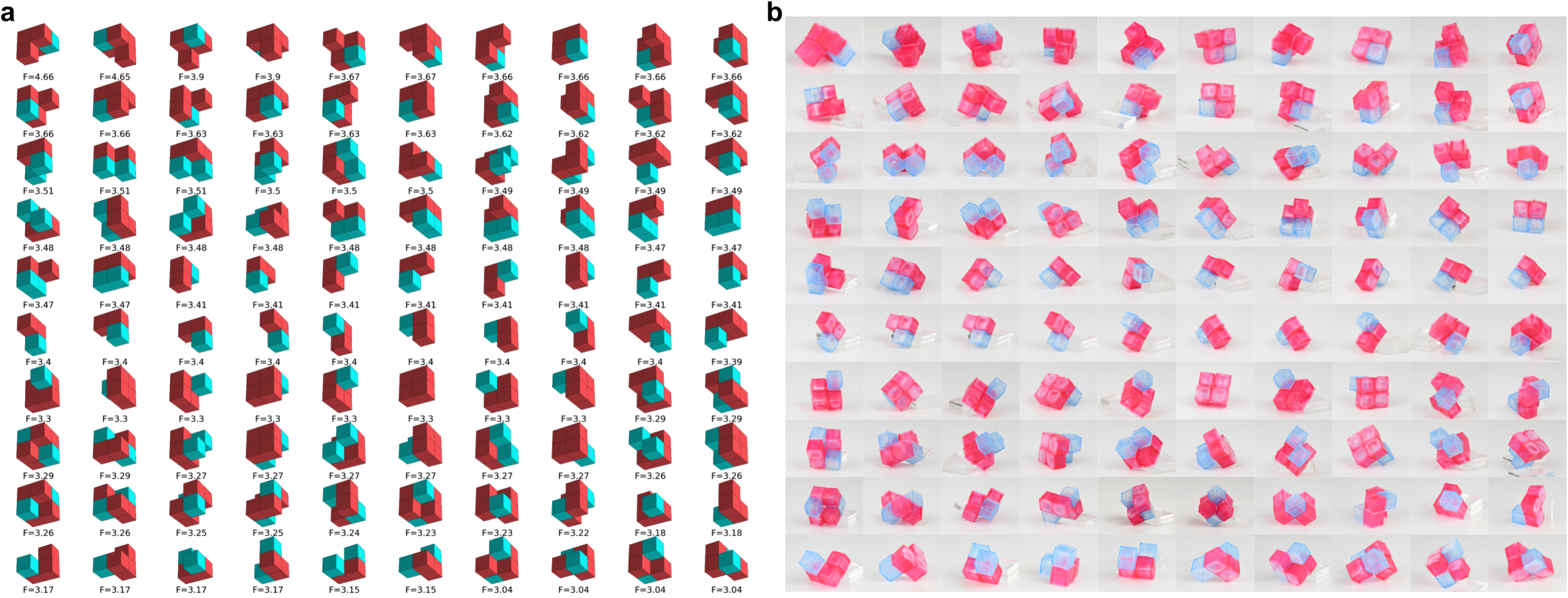}
\vspace{-18pt}
\caption{The top 100 simulated 2-by-2-by-2 configurations of passive (cyan) and volumetrically-actuating (red) voxels (\textbf{a}) were manufactured in reality (\textbf{b}).
} 
\label{fig:teaser}
\vspace{-24pt}
}

\maketitle

\begin{abstract}
The manual design of soft robots and their controllers is notoriously challenging, but it could be augmented|or, in some cases, entirely replaced|by automated design tools. 
Machine learning algorithms can automatically propose, test, and refine designs in simulation, and the most promising ones can then be manufactured in reality (sim2real).
However, it is currently not known how to guarantee that behavior generated in simulation can be preserved when deployed in reality.
Although many previous studies have devised training protocols that facilitate sim2real transfer of control polices, little to no work has investigated the simulation-reality gap as a function of morphology.
This is due in part to an overall lack of tools capable of systematically designing and rapidly manufacturing robots.
Here we introduce a low cost, open source, and modular soft robot design and construction kit, and use it to simulate, fabricate, and measure the simulation-reality gap of minimally complex yet soft, locomoting machines.
We prove the scalability of this approach by transferring an order of magnitude more robot designs from simulation to reality than any other method. 
The kit and its instructions can be found here:
\href{https://github.com/skriegman/sim2real4designs}{\color{blue}\tt\textbf{github.com/skriegman/sim2real4designs}}
\end{abstract}

\IEEEpeerreviewmaketitle

\section{Introduction}
\label{sec:intro}

The simulation-reality gap\footnote{Henceforth, ``the reality gap''|as coined by \citet{jakobi1995noise}.} for rigid-bodied robots is steadily closing.
Computational models of rigid body dynamics can now be regularized and tuned so that control policies optimized in simulation are just as successful when tested on the physical system \cite{bongard2006resilient,hwangbo2019learning}.
The reality gap for soft robots, on the other hand, remains uncharted.
It could be wider than the gap for rigid bodies, or not.
Soft bodies are more challenging to accurately simulate, design, and precisely control.
But, they are also, by definition, more permissive to simulation inaccuracies, design flaws, and control precision: 	
A soft gripper or foot will passively conform to complex objects and terrain, reducing the burden on the simulator to perfectly capture any single, ``true'' surface contact geometry.

Quantifying which soft robot designs, policies and behaviors can be faithfully simulated is critical not only for robotics, but also synthetic approaches to understand functional plasticity of biological systems during development and regeneration.
For both domains, testing candidate hypotheses in reality is expensive, time consuming, and, in some cases, dangerous.
With the recent development of several high-space, many-body, GPU-accelerated 
soft body simulators \cite{holden2019subspace,macklin2019non},
sim2real for soft robotics and synthetic biology has become more feasible.
However, because these simulators have yet to be employed to design physical systems, their transferability is currently unknown.

Previous work has demonstrated methods that promote successful sim2real transferal of soft object manipulation but not soft robot behavior.
For example, a rigid-bodied robot arm was successfully trained in simulation to fold towels and drape pieces of cloth over a hanger \cite{matas2018sim}.
However, the reality gap was not quantified beyond a binary success rate for each task.
Additionally, the robot's geometry was fixed and controllers were then optimized for it, whereas in the work reported here, the robot's geometry is part of the solution space.

\citet{hiller2011automatic} evolved the overall geometry and distribution of hard and soft materials in simulation, and transferred the structures and passive dynamics of various cantilever beams.
In a separate experiment that included actuating materials, Hiller and Lipson evolved the morphology and behavior of soft robots in simulation, and then built one of the evolved designs physically. 
However, in order to transfer the simulated behavior of this one design, the physical robot needed to be placed in a pressure and vacuum chamber,
whereas the hundreds of soft robot designs built here can be internally pressurized and actuated.

More recently, \citet{kriegman2019automated} subjected a simulated soft robot (composed of elastic voxels) to a series of damage scenarios that removed increasingly more of the robot's structure.
In each scenario, the robot was challenged to recover function (locomotion) by deforming its remnant structure, without changing its predamage control policy.
A pair of recovery strategies, automatically discovered by an evolutionary algorithm in simulation, were transferred to reality (using silicone ``voxels''), but function was not:
The physical system could deform its resting structure as dictated by the recovery strategy, but it could not locomote, before or after damage.
The physical robot was heavy, 
had high friction feet,
and was symmetrically actuated in phase, so it just oscillated in place.

To determine the particular challenges and opportunities of soft robot transferal, it would be useful to greatly scale up the number of design/policy pairs transferred. 
To this end, we present a soft robot design and construction kit based on the silicone voxel modules used in \cite{kriegman2019automated}, but miniaturized to increase stability, simplified to improve reproducibility, and arbitrarily actuated to permit the transferal of locomotion.

\begin{figure}[b]
    \vspace{-1em}
    \centering
    \includegraphics[width=\linewidth]{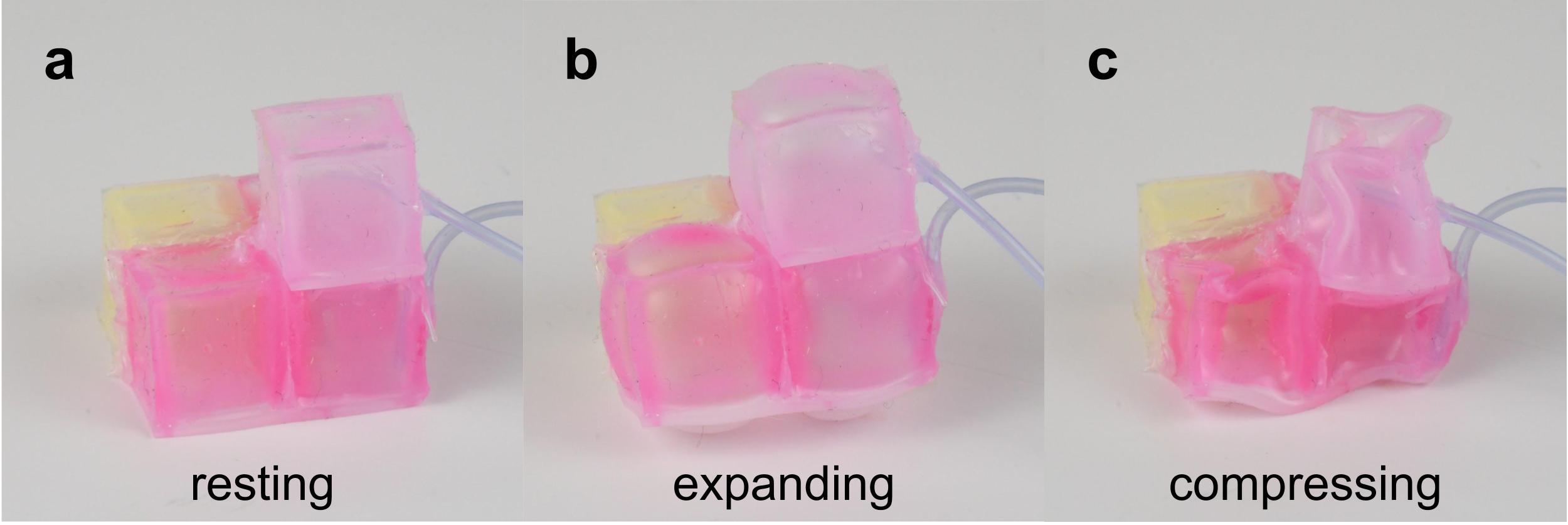}
    \vspace{-1.5em}
    \caption{A random morphology in the design space shown at atmospheric (resting; \textbf{a}), positive (expanding; \textbf{b}), and negative (compressing; \textbf{c}) pressure.}
    \label{fig:pressure}
\end{figure}

Other modular yet rigid-bodied robot design and construction kits exist, such as Molecubes \cite{zykov2007molecubes}.
However, our kit is easier, faster, cheaper, and safer to use.
In short, silicone is molded into hollow voxels, and tubing is attached to supply low pressure actuation from a hand pump,
causing volumetric changes in one or more of the voxels (Figs.~\ref{fig:pressure} and \ref{fig:real}).
For simple behaviors robust to actuation noise,
there is no need to use a highly-pressurized air supply or program microcontrollers for open-loop control.
There are also no expensive motors or power supplies.

Here, we employ the kit as an instrument to measure the reality gap as a function of morphology (Table~\ref{table:lit_review}).
To do so, we fabricated 108 morphologies (transferal of structure) and compared the behavior of nine simulated designs to their silicone equivalents (transferal of behavior).
We hope that the kit's affordability, safety, speed, and simplicity will generate increasingly more, and more reproducible, data about the automated design of increasingly competent soft machines.

\begin{table}[t]
\caption{\label{table:lit_review}Summary of published sim2real transference.}
\vspace{-1.25em}
\begin{center}
\begin{tabular}{lccc} 
    \toprule
    \textit{Author/citation} & \textit{Year} & \textit{Controllers} & \textit{Morphologies} \\ 
    \midrule
    
    \citet{Miglino1994Selection}    & 1994 & 1 & 1 \\  
    \citet{jakobi1995noise}         & 1995 & 2 & 1 \\  
    
    
    \citet{HARVEY1997205}           & 1997 & 4 & 1 \\  
    
    \citet{lipson2000automatic}     & 2000 & 3 & 3 \\
    
    \citet{bongard2006resilient}    & 2006 & 34 & 2 \\
    
    \citet{hiller2011automatic}     & 2011 & 1 & 5 \\ 
    
    \citet{koos2012transferability} & 2012 & 2 & 2 \\ 
    
    \citet{moeckel2013gait}         & 2013 & 1 & 1 \\  
    
    \citet{caluwaerts2014design}    & 2014 & 2 & 1 \\ 
    
    \citet{cully2015robots}         & 2015 & 10 & 10 \\
    
    \citet{cellucci20171d}          & 2017 & 1 & 3 \\
    
    \citet{tobin2017domain}         & 2017 & 1 & 1 \\ 
    
    
    \citet{rusu2017sim}             & 2017 & 1 & 1 \\ 
    
    \citet{peng2018sim}             & 2018 & 1 & 1 \\ 
    
    \citet{Pinto-RSS-18}            & 2018 & 3 & 1 \\ 
    
    \citet{tan2018sim}              & 2018 & 2 & 1 \\ 
    
    
    \citet{pmlr-v87-golemo18a}      & 2018 & 1 & 1 \\ 
    
    \citet{matas2018sim}            & 2018 & 3 & 1 \\ 
    
    
    
    
    \citet{kwiatkowski2019task}     & 2019 & 2 & 2 \\
    
    \citet{hwangbo2019learning}     & 2019 & 3 & 1 \\ 
    
    \citet{kriegman2019automated}   & 2019 & 1 & 5 \\
    
    \citet{nachum2019multi}         & 2019 & 3 & 1 \\  
    
    \citet{akkaya2019solving}       & 2019 & 1 & 1 \\
    
    \citet{rosser2019sim2real}      & 2019 & 1 & 16 \\
    
    \textbf{The results presented here} & \textbf{2019} & \textbf{1} & \textbf{108} \\
    \bottomrule
\end{tabular}
\end{center}
\vspace{-1.75em}
\end{table}

\section{Methods}
\label{sec:methods}


\subsection{The design space.}

Following \cite{hiller2011automatic} and \cite{kriegman2019automated}, our kit uses elastic voxels as building blocks of structure.
Here, we considered a \mbox{2-by-2-by-2} cartesian lattice workspace, within which voxels were connected together to form a robot.
At each x,y,z coordinate, voxels could either be passive, volumetrically actuated, or absent, yielding a total of $3^8=6561$ different configurations. 
We evaluated each configuration in simulation.

\begin{figure}[t]
    \centering
    \includegraphics[width=\linewidth]{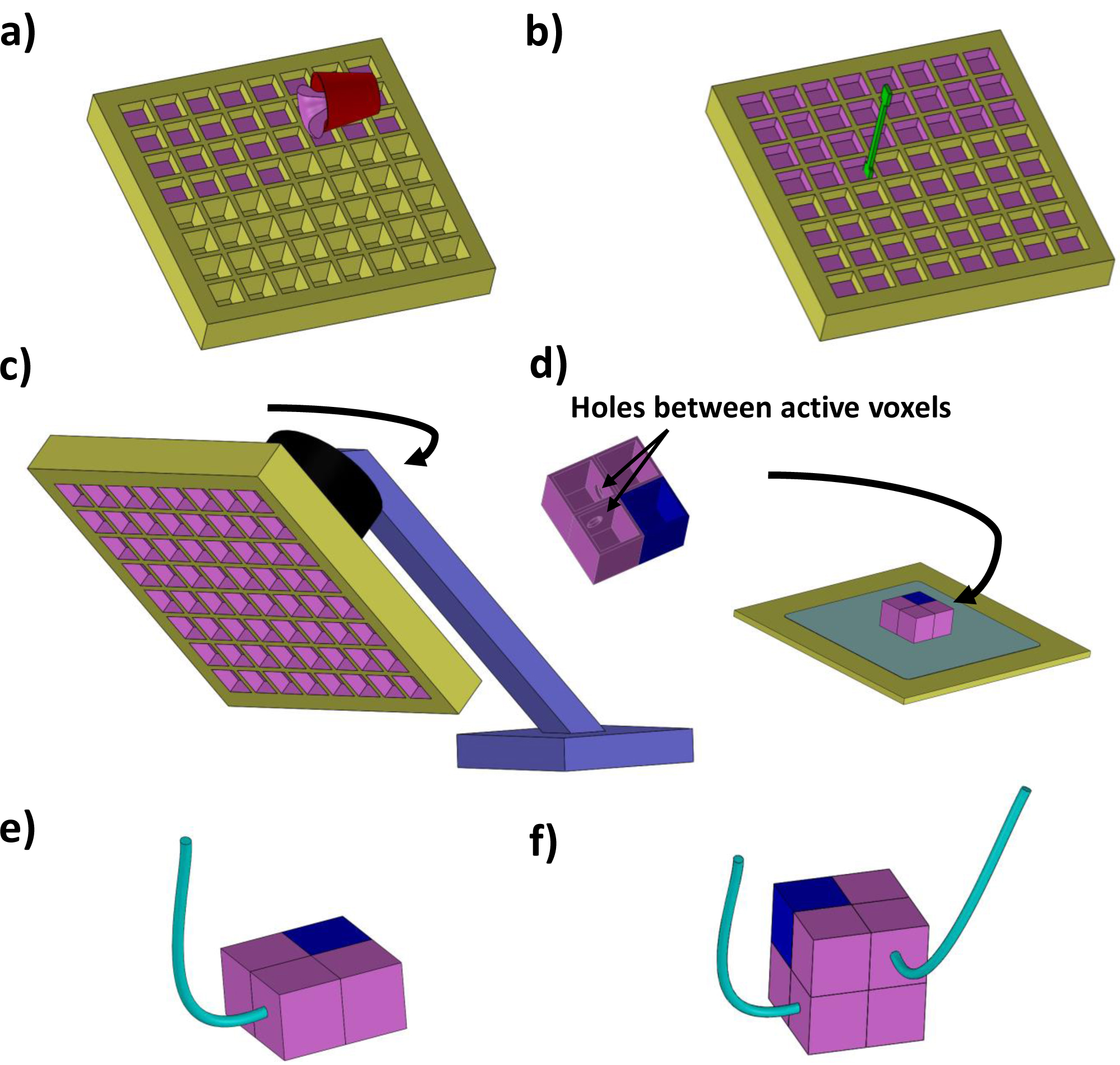}
    \caption{\textbf{Manufacturing modular soft robots.}
    Hollow, silicone voxels were created by partially filling an open-face mold with silicone (\textbf{a}), 
    using a spatula to spread it along the interior walls (\textbf{b}), and then securing the mold to a 1-axis rotational molding machine (\textbf{c}). This process allowed excess silicone to drip out of the mold, while spreading the remaining silicone into a thin uniform layer.
    The cured, bottomless voxels were then appropriately arranged and connected for each x,y slice of the design, and bonded with a shared bottom layer (\textbf{d}).
    Finally, tubing was attached (\textbf{e}), and the slices were stacked and bonded to form the design (\textbf{f}). 
    Video:
\href{https://youtu.be/jbQ2T7jIYRU}{\color{blue}\tt\textbf{youtu.be/jbQ2T7jIYRU}}.
    }
    \label{fig:real}
    \vspace{-1.5em}
\end{figure}

\subsection{The simulation.}

We used the soft-body physics engine Voxelyze \cite{hiller2014dynamic} to simulate robots composed of actuating and/or passive, elastic voxels.
The simulator models the distance between adjacent voxels as Euler-Bernoulli beams (critically damped; $\zeta=1$).
Additionally, a collision detection system monitors the distance between the voxels on the surface of the robot at each timestep.
If a pair of surface voxels are detected to collide (intersect), a temporary beam (underdamped; $\zeta=0.8$) is constructed between the two until the collision is resolved. 

Designs were simulated with a gravitational acceleration of -9.81 m/s$^2$, and initialized on top of an infinite surface plane at $z=0$.
Coulomb friction is applied to voxels in contact with the surface plane.
Voxels were simulated to have 1 cm$^3$ resting volume (resting beam lengths), with Young's modulus $10^7$ Pa, Poisson's ratio 0.35, 
and coefficients of static and kinetic friction of 1 and 0.5, respectively.
These hyperparameters were adopted from \cite{kriegman2019automated}.
For more details about how the physics are actually modeled, see \cite{hiller2014dynamic}.

Volumetric actuation was implemented by varying the rest length between voxels, in all three dimensions, when computing the elastic force between them.
Volumetric expansion in simulation and reality are both roughly spherical
(Fig.~\ref{fig:pressure}b), but compression in reality is more complex
and difficult to simulate: 
the voxels buckle (Fig.~\ref{fig:pressure}c).
So volumetric actuation was here limited to expansion only ($+90\%$ rest volume).
The active voxels expand in phase with each other 
as dictated by
a central pattern generator:
a sine wave with frequency 4 Hz and amplitude 1.9 cm$^3$.
When the sine wave is at or below zero, the active voxels remain at their resting volume (1 cm$^3$).
This produced quasistatic dynamics.

Each design was simulated for 8831 timesteps, with a stepsize of 0.000453 seconds, resulting in a total simulation time of 4 seconds.
During the first 552 times steps (0.25 sec), the design was allowed to settle under gravity before actuation begins, ensuring that movement (if any) is a result of actuation, rather than passively falling forward.
Just before actuation, the design's initial center of mass is recorded as $(x_0, y_0, z_0)$.
The active voxels are then actuated for 3.75 sec at 4 Hz, or 15 actuation cycles.

An exhaustive search of all 6561 designs (in batches of 50) took 58 CPU hours (1.8 wall-clock hours) on a single AMD Ryzen threadripper 1950X 16-core/32-thread processor.
Fitness was taken to be the net displacement (away from the origin in any direction) of the design's center of mass, in terms of euclidean distance in the plane,
where the origin is defined by the $x,y$ components of the design's initial center of mass $(x_0, y_0)$.
Fitness is thus defined as:
\begin{equation}
    \label{eq:fitness}
    F=\sqrt{(x_t-x_0)^2+(y_t-y_0)^2} \,,
\end{equation}
where $x_t,y_t$ are the final coordinates of the design at the end of the evaluation period.

\subsection{Reality.}

Following \citet{kriegman2019automated}, simulated voxels were realized physically as pneumatically-actuated, hollow silicone voxels.
The physical robot in \cite{kriegman2019automated} was constructed to transfer symmetrical shape change, so its actuated voxels were distributed symmetrically and hooked into a single pressure inlet.
Thus, pressure oscillations occurred symmetrically in phase, and the robot could only pulse in place.
Moreover, due to thin voxel walls relative to overall voxel size, and the tubing and glue used to bond them together, the robot in \cite{kriegman2019automated} could not fully support its own weight.
The robot was lifted off the ground by placing it on top of a small petri dish, positioned underneath a segment of entirely passive voxels in the center of the robot's ventral surface.
This permitted ventral (and more extreme global) changes in surface curvature, yielding successful sim2real transfer of shape change, 
but not locomotion.

The construction kit presented here rectifies the weight issue by miniaturizing the voxels|voxel length was halved (from 3cm to 1.5cm) and the wall thickness remained the same (1mm), reducing voxel mass from 4.3g to 1.2g (including tubing but not pneumatic connectors).
Further, the inter-voxel tubing and glue was replaced with holes punched through the walls of adjacent active voxels in the same x,y slice, before attaching them with a shared bottom layer (Fig.~\ref{fig:real}d).
Finally, locomotion is now possible because separate contiguous sections of voxels in each slice can be arbitrarily actuated in or out of phase with other sections across the body.

\subsection{The build protocol.}

The voxels were manufactured using a single-axis rotational molding machine.\footnote{The required materials are listed at the end of the manuscript in Table~\ref{table:core_materials}.}
First, an open-face mold was fabricated by interlacing 26 acrylic strips into a flat base, to form a lattice of cubic concavities, resembling an ice-cube tray (Fig.~\ref{fig:real}a). Mold components were laser-cut (VLS2.30, Universal Laser System) from a flat acrylic sheet with a thickness of 0.025 inch.
Next, silicone (Dragon Skin 10 Fast; Smooth-On, Inc.) was poured into the acrylic mold (Fig.~\ref{fig:real}a), and a spatula was used to spread the silicone along the interior walls of each cavity (Fig.~\ref{fig:real}b). Colored pigment was added to each batch of silicone to indicate whether the voxel was active or passive, simplifying the assembly process. Here we used pink for active voxels and blue or yellow for passive voxels.

The mold was then flipped upside down and secured to a 1-axis rotational molding machine. The machine was clamped to a table with binder clips, angled $45^{\circ}$ relative to horizontal, and set to rotate $90^{\circ}$ every 45 seconds (Fig.~\ref{fig:real}c).
This allowed the silicone to flow and evenly coat the walls of the mold, as excess silicone dripped out. 
After the voxels partially cured for 25 minutes at room temperature, the mold was moved to an incubator, with a temperature of $60^{\circ}$C for another 20 minutes. 
(Without an incubator, the silicone will take 75 minutes to fully cure at room temperature.)

The above steps were then repeated to add an additional layer of silicone.
Once the second layer cured, the bottomless voxels were removed from the mold using an X-Acto knife, and excess silicone around their edges was trimmed.

In the next step, each x,y slice (or dorsal plane) of the design was assembled by using Sil-Poxy (Smooth-On, Inc.) to bond adjacent voxels and prevent the slice from shifting.
Holes were then punched between adjacent active voxels so that contiguous collections of voxels could be actuated together in phase.
Each actuator group needed to contain at least one voxel on the surface of the design so that it could be controlled by an external pressure inlet.
To create the bottom layer, two 1mm-thick rulers were attached to an acrylic substrate using double-sided tape and silicone was poured in the space between them. 
Then, the slice of bottomless voxels was flipped, open-side down, onto this uncured silicone layer (Fig.~\ref{fig:real}d). 

After the bottom layer cured, a thin layer of silicone was applied with a popsicle stick along the outermost portions of the interstices of the voxels, bonding adjacent voxels (without gluing over inter-voxel holes).
Then, the slice was cut from the silicone sheet and a hole was poked into the side of one exterior voxel from each group of active voxels.
Next, a 1/32'' ID silicone tube was inserted into the hole, and glued in place with Sil-Poxy, applied with a Q-tip (Fig.~\ref{fig:real}e). 
The end of this tube was then connected to a straight pneumatic connector, which was connected to 1/16'' ID silicone tubing.

Occasional imperfections in alignment, silicone thickness, or inter-voxel hole sizes would result in leaky structures. 
Leaks were detected by filling a beaker with water, submerging the voxels, and inflating them. 
Bubbles would emanate from leaks, which were repaired with Sil-Poxy.
After repairing any leaks, the slices were stacked on top of each other and bonded together using a thin layer of silicone (Fig.~\ref{fig:real}e). 
Finally, these layers were connected pneumatically with assorted pneumatic connectors, attached to 1/16'' ID silicone tubes.

\section{Results}
\label{sec:results}

To test the effects of morphology on fitness and sim2real transfer success, it is useful to first visualize the design space. 
\begin{wrapfigure}{r}{0.465\linewidth}
\vspace{-1em}
\includegraphics[width=\linewidth]{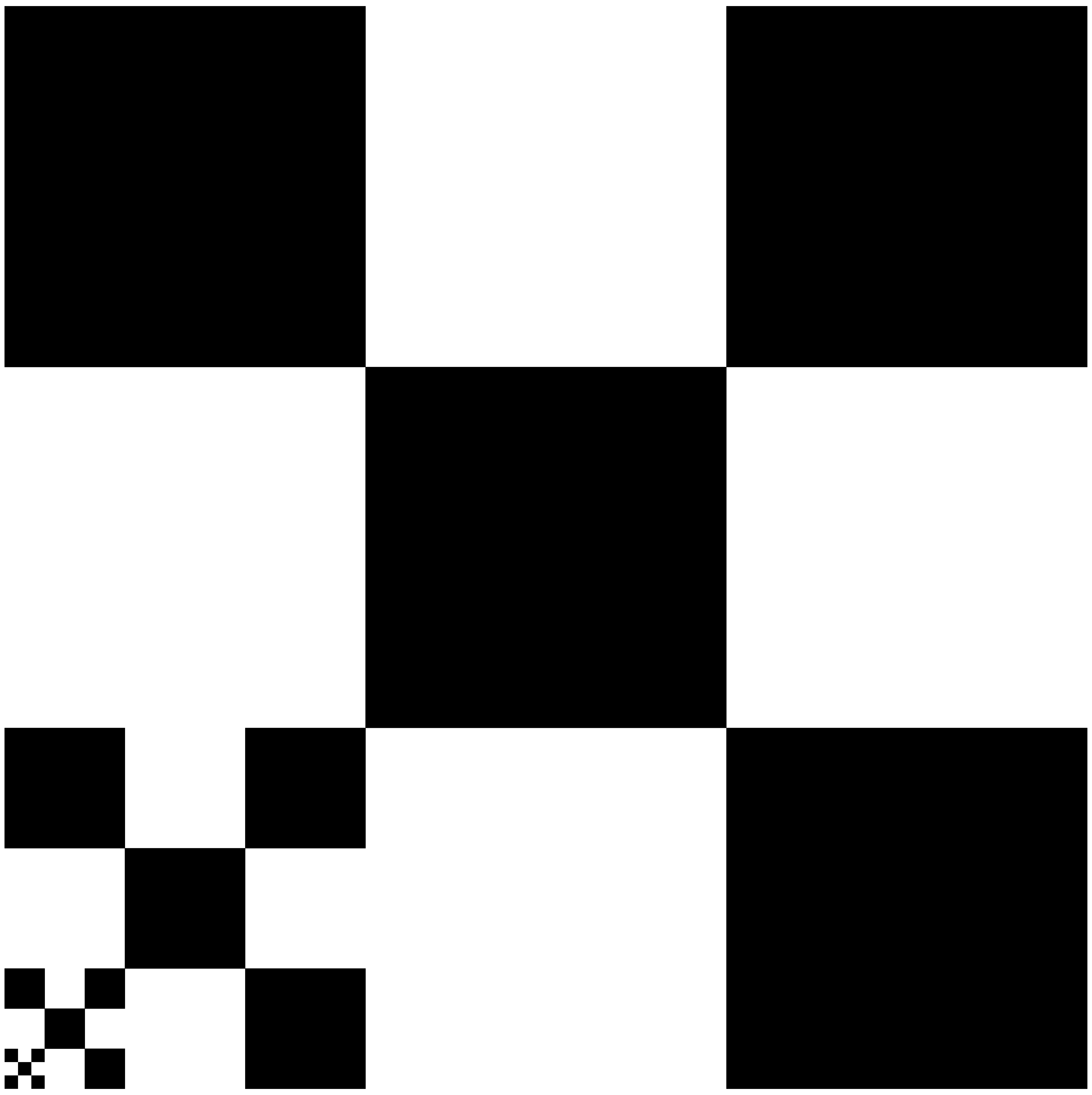}
\vspace{-18pt}
\caption{The 2D tessellation of 8D ternary vector space used in Fig.~\ref{fig:sim}.}
\label{fig:tesselation}
\vspace{-0.5em}
\end{wrapfigure}
However, because there are eight cartesian voxel coordinates in the chosen workspace, the design space here is eight dimensional, which is difficult to draw (let alone conceptualize) without dimensionality reduction.
By nesting the dimensions of a search space onto a single plot (Fig.~\ref{fig:tesselation}), the entire space can be visualized as a 2D heatmap.
This strategy was used by \citet{cully2015robots} to neatly visualize the predicted fitness of a very large library of control policies, as a function of the time a robot's six limbs were in contact with the simulated ground plane: 6D quinary control space was mapped to 2D, by nesting pairs of dimensions within each other.

\begin{figure}[!b]
    \vspace{-2em}
    \centering
    \includegraphics[width=\linewidth]{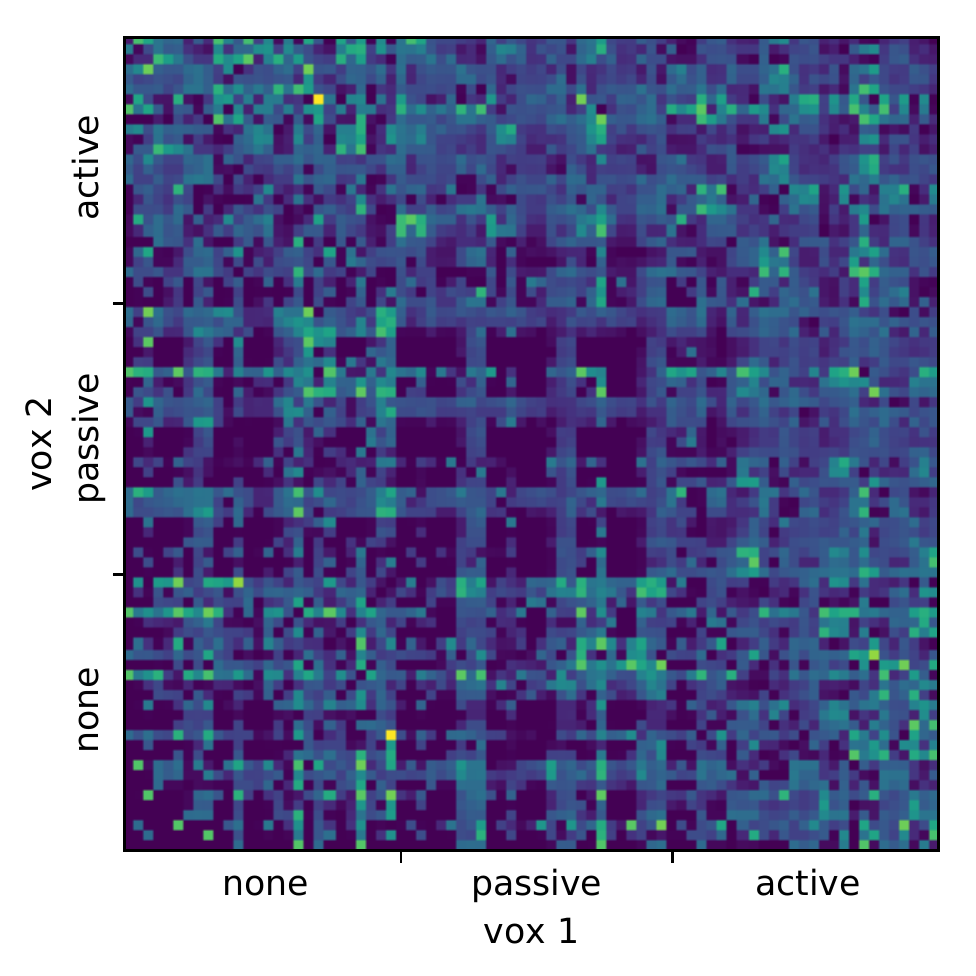}
    \vspace{-2em}
    \caption{\textbf{Simulating modular soft robots.}
    The design space is plotted as a heatmap, containing one cell for each of the 6561 possible configurations.
    Lighter colored cells are fitter designs (Eq.~\ref{eq:fitness}).
    Each design is defined by a vector of eight ternary values, indicating what kind of voxel (none, passive, or active) the design contains at the eight lattice points in the $2\times2\times2$ workspace.
    The 8D ternary vector is reduced to a 2D heatmap by nesting pairs of dimensions within each other: four, nested $3\times3$ grids result in a $3^4\times3^4=81\times81$ overall heatmap.
    }
    \label{fig:sim}
\end{figure}

\begin{figure*}[t]
    \centering
    \includegraphics[width=\linewidth]{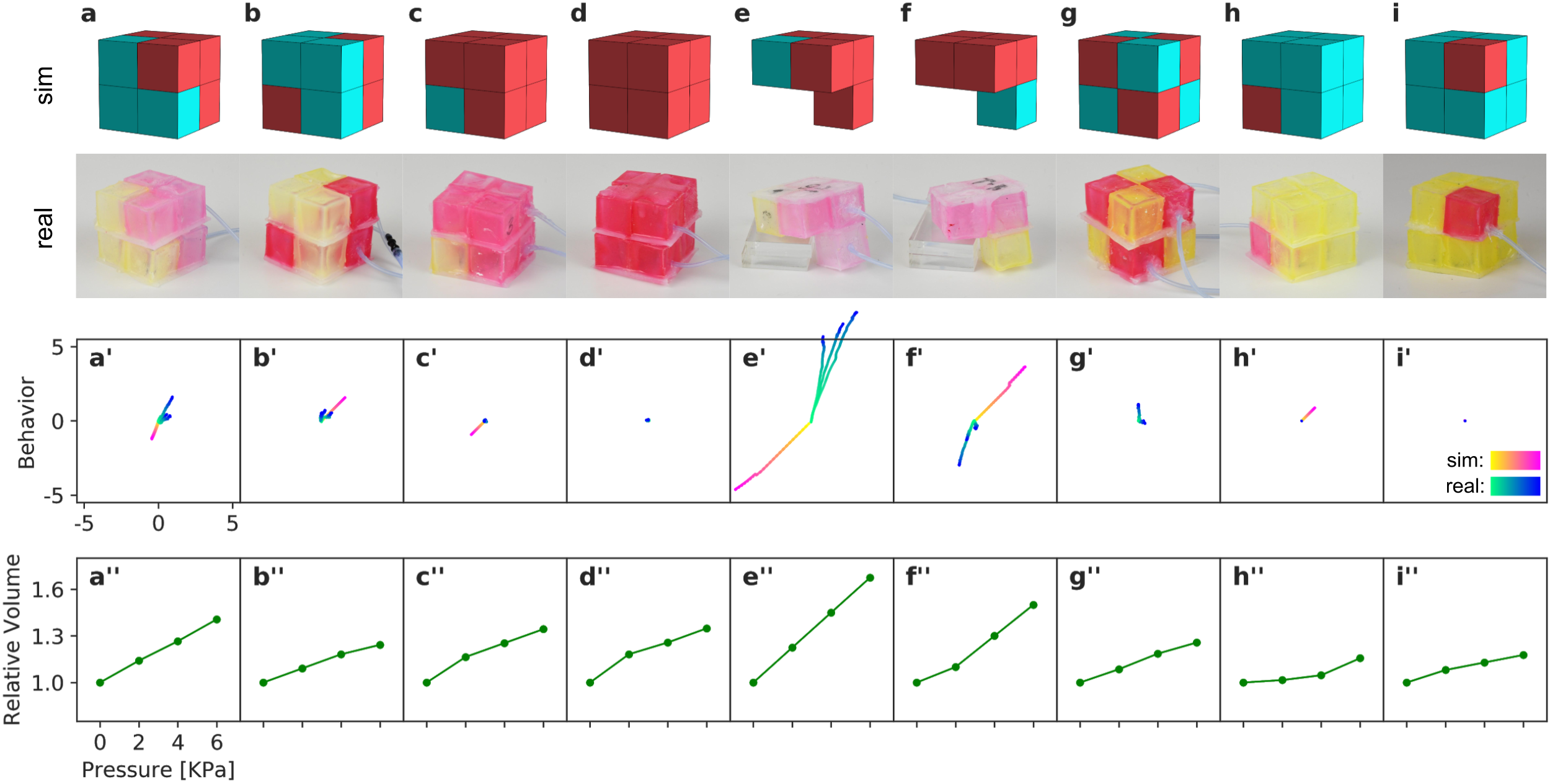}
    \vspace{-1.75em}
    \caption{\textbf{Measuring transferal from simulation to reality}.
    Nine designs (\textbf{a-i}) were evaluated three times each in reality (green-to-blue gradient colored curves in \textbf{a\textquotesingle-i\textquotesingle}).
    The behavioral trajectories start at the origin (green) and end at the robot's final XY destination (blue) (in centimeters).
    The simulated movement tracks (yellow-to-pink curves) are superimposed on top of the real ones.
    The relative volume (normalized by rest volume) was also recorded for each design at four points during actuation under water (\textbf{a\textquotesingle\textquotesingle-i\textquotesingle\textquotesingle}).
    The simulated and real behavior of designs e and f can be observed here:
    \href{https://youtu.be/UqjvmkYa9u4}{\color{blue}\tt\textbf{youtu.be/UqjvmkYa9u4}}.
    }
    \label{fig:transfer}
    \vspace{-1.25em}
\end{figure*}

Here, the 8D ternary morphology space was reduced to 2D by plotting pairs of dimensions nested within each other (Fig.~\ref{fig:sim}).
The pixel in the exact center of Fig.~\ref{fig:sim}, for instance, represents the configuration consisting entirely of passive voxels, and thus cannot locomote ($F=0$).
Likewise, the pixel in the top right-hand corner of the heatmap represents the configuration of all active voxels (Fig.~\ref{fig:transfer}d), which actuated symmetrically in phase, and thus (given its flat ventral surface) could not locomote across the flat ground plane ($F=0$).
Finally, the pixel in the bottom left-hand corner contains no voxels at all, and thus $F=0$.

For locomotion, a good design obviously needs to have a body, rather than none at all.
With open-loop, in-phase actuation, designs also need to have asymmetrical mass and/or actuator distributions, or they will not generate any forward movement.
However it is not clear, even for this minimal design space, exactly which asymmetrical designs will yield the highest fitness.
Yet we can see small clusters and lines of similarly colored pixels in Fig.~\ref{fig:sim}, representing morphologically similar designs with similar fitness.
This suggests that these configurations and substructures would be relatively stable under random mutations or errors in fabrication.

Because fitness was measured by displacement in any direction away from the origin (Eq.~\ref{eq:fitness}), there are four configurations|rotations, in the x,y plane, of a single geometry and distribution of passive and active voxels|with different behaviors (they moved in different directions) but very similar (if not identical) fitness.
There were also some configurations that, when rotated upward (in the x,z or y,z plane) fell into the same basic orientation and behavior but with a slightly different heading.
Thus, configurations with similar fitness (similarly colored pixels) are reflected across multiple, nested planes of symmetry in Fig.~\ref{fig:sim}.
These symmetries can also be seen in the manufactured robots (Fig.~\ref{fig:teaser}b).
The uniqueness of designs (i.e., the size of the search space of morphologies) is therefore a function of how behavior is measured.

Fig.~\ref{fig:transfer} shows the behavior of nine different designs in simulation and reality.
The real robot was actuated 90 times at 6 kPa pressure on a surface covered with cornstarch 
(Argo\textregistered, ACH Food Companies, Inc.) 
to reduce friction, and is compared to 23 simulated actuation cycles.
Seven of the nine designs filled the cubic workspace with passive and active voxels, while the other two share a more complex geometry: a single-voxel limb attached to the face of a 2-by-2 plane of voxels (Fig.~\ref{fig:transfer}e,f).
In one, the limb is active (Fig.~\ref{fig:transfer}e), in the other it is passive (Fig.~\ref{fig:transfer}f).
These two designs achieved the two highest fitness scores (Eq.~\ref{eq:fitness}), in both simulation and reality.

By this measure, the reality gap appears small.
However, these simulated designs move very differently from their manufactured equivalents.
The simulated morphology in Fig.~\ref{fig:transfer}e pushes off its active limb, whereas in reality the design uses its limb to pull itself forward, in the opposite direction.
Likewise, the simulated morphology in Fig.~\ref{fig:transfer}f pushes off its active 2-by-2 torso, whereas in reality the design uses its torso to pull itself forward, in the opposite direction.

\citet{majidi2013influence} showed that the interfacial shear strength and coefficient of friction of the surface on which their soft robot undulated determined the direction of locomotion. They decomposed friction into load- and area controlled terms for point and surface contacts, respectively. 
On slippery surfaces with low interfacial shear resistance, the robot anchored about the point contact (expanded section) for locomotion and pulled its surface contact (passive segment). However, on surfaces with high interfacial shear resistance, the robot anchored about the surface contact and pulled the point contact toward it. 
We hypothesize that such differences in tribological properties could have caused our designs to move in opposite directions in simulation and reality.

\begin{figure}[t]
    \centering
    \includegraphics[width=0.95\linewidth]{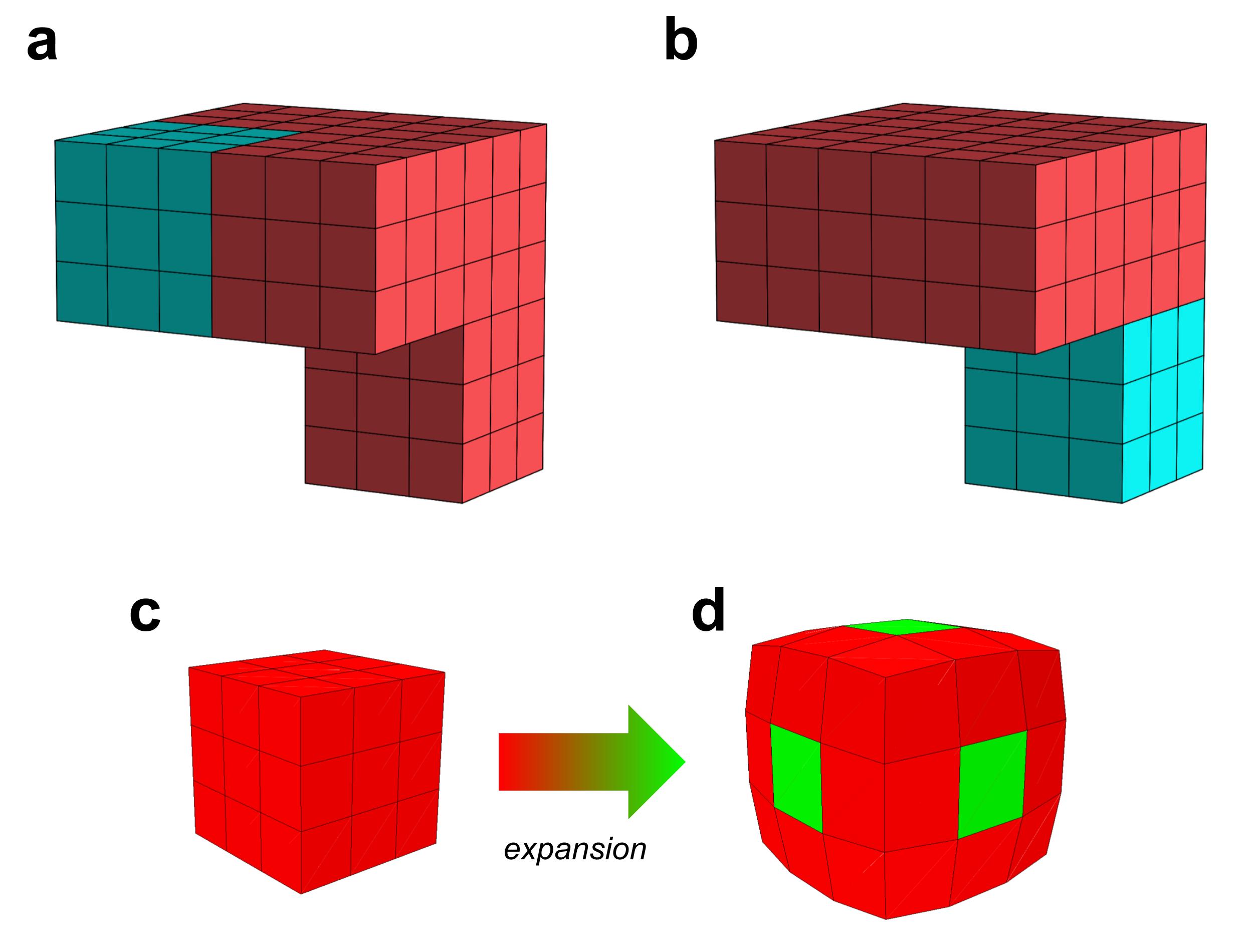}
    \vspace{-1em}
    \caption{A higher resolution model in which each silicone voxel is approximated by a 3-by-3-by-3 group of simulated subvoxels: a high-res voxel.
    The design in \textbf{a} and \textbf{b} are high-res instantiations of those in Figs.~\ref{fig:transfer}e and \ref{fig:transfer}f, respectively. 
    Spherical volumetric expansion in a high-res voxel (\textbf{c}) was approximated by increasing the rest length between the centermost subvoxel and the subvoxels at center of each face (green subvoxels in \textbf{d}).
    }
    \label{fig:hi_res_sim}
    \vspace{-1.25em}
\end{figure}

In an attempt to test this hypothesis and reduce the simulation-reality discrepancies that cause the virtual configurations in Fig.~\ref{fig:transfer} to move differently than their physical realizations, we performed a grid search of various simulation hyperparameters, including the coefficients of static and kinetic friction.
However, we could not identify a pair of friction coefficients that resulted in correct movement heading for all nine of the behaving designs (Fig.~\ref{fig:transfer}a\textquotesingle-i\textquotesingle).
This could be due to either low precision or low accuracy of the model.
To isolate and test the former possibility, we increased the resolution of the simulated surface contact geometry by modeling each silicone voxel as a 3-by-3-by-3 group of simulated ``subvoxels'' (Fig.~\ref{fig:hi_res_sim}),
and then re-ran the parameter sweep.
Still, we could not find friction settings in which the simulated movement direction matched the ground truth across all designs simultaneously.
This suggests that the accuracy of Coulomb friction model may be insufficient to model this type of movement.

The Coulomb approximation assumes that friction is simply proportional to the vertical component $N$ of the reaction force, and independent of the contact area. 
However, friction is also a function of the surface area and interfacial shear strength $\tau$, a fixed constant which is mostly governed by adhesion or mechanical interlocking between the contacting surfaces. 
A better model would thus consider friction as a function of both the normal force and the interfacial shear strength.
However, before fundamentally changing the simulator,
we plan to evaluate designs in noisy environments with imperfect control over actuation characteristics to avoid ascribing high fitness to designs that exploited unrealistic properties of the simulation \cite{jakobi1995noise}.
Additionally, data from reality could be used to automatically tune  the geometry and resolution of the simulated finite elements \cite{bongard2006resilient}, or to predict the kinds of behaviors that are more likely to successfully transfer \cite{koos2012transferability}, and which should be tested next \cite{bongard2006resilient}. 
Concurrently, we are investigating additional physical surfaces with varied tribological properties in an attempt to match reality to simulation.



\section{Discussion}
\label{sec:discussion}

In this paper, we introduced a low cost, open source platform for designing and rapidly building soft robots, and used it to transfer 
108 different morphologies (voxels on a cartesian grid) from simulation to reality.
We then measured the reality gap as function of the robot's design (geometry and distribution of passive and actuating voxels) by tracking the behavior of nine transferred morphologies.
Under one measure (net displacement) the reality gap appeared rather small, but under another (velocity) the gap was much wider, likely due to oversimplified tribological contacts between the simulated ground plane and the robot's ventral surface \cite{majidi2013influence}.

Although most of the transferred designs (99 out of the 108) were not actuated in reality, they nevertheless served an important function:
they were \textit{sketches}.
Sketches let us think more clearly about the behavior or properties (e.g., stability) of a design without investing the additional time and resources required to fully build and examine the design itself.
Sketches, in other words, greatly increase the breadth of exploration in design space.
All sim2real methods embrace this utility of simplifying sketches.
Simulation, after all, is also a sketch.

However, there is a tacit assumption in robotics about Depth First.
A typical sim2real experiment begins by sending a
complicated robot design across the reality gap,
and then endeavors to learn transferable policies that control the morphology in all its complexity.
But this is not how most design proceeds.
An architect first roughly sketches a structure, say, a bridge, on the back of a napkin.
A diversity of designs are then generated, tweaked, discarded or provisionally accepted|at this shallow level of napkin realism|before more detailed blueprints are drawn under more stringent constraints.
Blueprints, too, undergo this breadthwise evolution, before the most promising are realized physically, first as scale models (built from matchsticks and glue instead of concrete and steel), then, finally, at full scale and cost.
This incrementally weeds out nontransferable features and adds mechanical complexity only when and where it is necessary to do so, rather than globally from the outset.

The assumption that the reality gap can be bridged by policy search alone, with a single robot design, is groundless. 
The desired behavior of a robot is typically much more complicated than that of architecture.
This suggests the necessity of more, not less, sketches.
Soft robots are more complicated still.
This makes their automated design that much more appealing, but implies the need for even greater breadth of sketches, at more intermediate levels of realism.
Though not every experiment will need to start from a blank slate.
Instead, designers (whether human or AI) could leverage prior knowledge to reject truly awful designs before sketching them in the first place.
The designs transferred here add to a growing database (prior probabilities) about which and how well different soft robot designs and behaviors can be realized physically.
Our construction kit has the potential to further increase this data by lowering not only cost and build times but also the barrier of entry to soft robotics for non-experts.

The generality of such data beyond robotics is currently not known, but it could also have important implications for developmental biology and regenerative medicine.
The bioelectric and genetic control policies that orchestrate adaptive remodeling of growth and form in organisms are not yet understood, but could, in future, be reverse-engineered by machine learning, and then controlled externally to induce regeneration in otherwise non-regenerative organisms (such as adult humans), or to reprogram otherwise unbounded cancerous growth toward functional organogenesis \cite{levin2013reprogramming}.
However, such advances in regenerative medicine and synthetic morphology will only be possible if hypotheses generated in simulation are transferable and testable in reality.

\section*{Acknowledgements}

This work was supported by 
NSF award EFRI-1830870 
and
DARPA contract HR0011-18-2-0022.

\renewcommand{\arraystretch}{1.3}
\setlength\tabcolsep{5pt}

\begin{table*}
\caption{\label{table:core_materials}The construction kit for making fifty 2-by-2-by-2 designs.}
\vspace{-1.5em}
\begin{scriptsize}
\begin{center}
\begin{tabular}{clcl} 
 \toprule
 \textbf{No.} & \textbf{Part} & \textbf{Cost} & \textbf{Notes} \\ 
 \toprule
 1 & Dragon Skin 10 FAST, 2 lbs kit (Smooth-On, Inc.) & \$32.21 & Created voxel bodies and connected x,y slices. \\ 
 \rowcolor{gray!25}
 2 & Sil-Poxy, 3 ounce tube (Smooth-On, Inc.) & \$30.72 & Secured tubing and repaired air leaks in six-sided voxels. \\ 
 3 & 1/32'' ID silicone tubing, 52 ft & \$34.32 & 50A shore hardness; connected voxels to actuation system. \\ 
 \rowcolor{gray!25}
 4 & 1/16'' pneumatic plastic connectors, 200 pieces & \$100 & Straight and T shaped. \\ 
 5 & 1/16'' ID silicone tubing, 65.6 ft & \$62.98 & 50A shore hardness; part of actuation system. \\ 
 \rowcolor{gray!25}
 6 & 60 mL Plastic Syringe (McMaster–Carr, Inc.) & \$3.13 &  Used to hand-actuate voxels while checking for leaks. \\
 \multirow{2}{*}{7} & \multirow{2}{22em}{1000 mL beaker (PYREX™ VISTA™ Griffin, Fisher Scientific International, Inc.)} & \multirow{2}{*}{\$14} & \multirow{2}{*}{Voxels were inflated in this beaker filled with water to detect leaks.} \\
  & & & \\ 
  \rowcolor{gray!25}
 8 & Hole punch 1/4 rectangle (Fiskars) & \$10.35 & Created holes between voxels within a slice so they actuated as a unit. \\ 
 9 & X-Acto Knife (McMaster–Carr, Inc.) & \$4.11 & Cut bottomless voxels out of acrylic mold, and trim the edges. \\ 
 \rowcolor{gray!25}
 10 & 100 mL mixing cups (VWR international), pack of 100 & \$66 & Where silicone was mixed. \\ 
 11 & Spatula (McMaster–Carr, Inc.) & \$7.50 & Spread the Dragon Skin on the edges of the acrylic mold. \\ 
 \rowcolor{gray!25}
 12 & 2 Scrap acrylic sheets, 12''$\times$12''$\times$1/8'' & \$10.58 & Collected scrap Dragon Skin; the surface where voxel bottom were created. \\ 
 13 & 2-1mm thick 30 cm metal rulers & \$6.99 & Used to set a thickness for sixth voxel side (bottom layer). \\ 
 \rowcolor{gray!25}
 14 & Double-sided tape & \$4.99 & Adhered the rulers to a scrap acrylic sheet. \\ 
 15 & Simple 30 cm metal ruler & \$3.99 & Spread a thin sheet of Dragon Skin onto scrap acrylic sheet for sixth side. \\ 
 \rowcolor{gray!25}
 16 & Popsicle stick (11.3 cm $\times$ 1 cm), box of 1000 & \$13.49 & Applied thin layer of silicone to bond adjacent voxels and x,y slices. \\ 
  17 & Cotton-tipped applicators, 6 inch, box of 1000 & \$8.99 & (McKesson Corp.) Used to spread Sil-Poxy. \\
  \rowcolor{gray!25}
 18 & Disposable Gloves (Halyard Inc.), box of 100 & \$8.95 & Wore when handling uncured silicone. \\ 
 19 & 2 Acrylic sheets, 12''$\times$12''$\times$1/8'' (McMaster–Carr, Inc.) & \$18.30 & Used to manufacture the open-face acrylic mold with laser-cut. \\ 
 \rowcolor{gray!25}
 20 & 41 mm binder clips, pack of 12 & \$7.99 & Held the acrylic mold onto the rotational molding machine. \\
 \toprule
  & \textbf{The rotational molding machine:} & & \\
 \toprule
 \rowcolor{gray!25}
 21 & Acrylic sheet, 12''$\times$12''$\times$1/4'' (McMaster–Carr, Inc.) & \$17.34 &  2$\times$ triangular plates, 3$\times$ motor mount - supports for rotational machine. \\ 
 22 & Acrylic sheet, 12''$\times$12''$\times$1/4'' (McMaster–Carr, Inc.) & \$17.34 & Mounting plate; holes were cut to minimize weight. \\ 
 \rowcolor{gray!25}
 23 & Pololu 4756 DC rotational motor & \$39.95 & Used for rotational molding machine. \\ 
 24 & Pololu 1999 mounting hum  & \$7.95 & Used to mount rotational molding machine. \\ 
 \rowcolor{gray!25}
 25 & Arduino Uno Microcontroller & \$22 & Controlled rotation timing and degree. \\ 
 25 & Arduino Motor Shield  & \$19.95 & Controlled rotation motor. \\  
 \rowcolor{gray!25}
 26 & AC charger & \$10.42 & 12mm$\times$2.1$\times$5.5mm barrel jack, 12V, Supplies power to the Arduino. \\ 
 27 & 8020 T-Slotted Solid 1'' beams (McMaster-Carr, Inc.) & \$12.31 & 2$\times$10 cm, 2$\times$40 cm, Supports for rotational molding machine. \\ 
 \rowcolor{gray!25}
 28 & 8020 screws and T-nuts & \$7.92 & Connected 8020 beams. \\ 
 29 & ``M3$\times$10'' screws & \$1.24 & Used to mount rotational molding machine. \\ 
 \rowcolor{gray!25}
 30 & Irwin QuickGrip 12''$\times$2.75'' Clamp & \$15.99 & Held rotational molding machine to table. \\
 \toprule
  & \textbf{Grand total:} & \textbf{\$622} &  \\
 \bottomrule
\end{tabular}
\end{center}
\end{scriptsize}
\vspace{-2.5em}
\end{table*}

\bibliographystyle{plainnat}
\bibliography{main}

\begin{thebibliography}{30}
\providecommand{\natexlab}[1]{#1}
\providecommand{\url}[1]{\texttt{#1}}
\expandafter\ifx\csname urlstyle\endcsname\relax
  \providecommand{\doi}[1]{doi: #1}\else
  \providecommand{\doi}{doi: \begingroup \urlstyle{rm}\Url}\fi

\bibitem[Akkaya et~al.(2019)Akkaya, Andrychowicz, Chociej, Litwin, McGrew,
  Petron, Paino, Plappert, Powell, Ribas, et~al.]{akkaya2019solving}
Ilge Akkaya, Marcin Andrychowicz, Maciek Chociej, Mateusz Litwin, Bob McGrew,
  Arthur Petron, Alex Paino, Matthias Plappert, Glenn Powell, Raphael Ribas,
  et~al.
\newblock Solving rubik's cube with a robot hand.
\newblock \emph{arXiv preprint}, 2019.
\newblock URL \url{https://arxiv.org/abs/1910.07113}.

\bibitem[Bongard et~al.(2006)Bongard, Zykov, and Lipson]{bongard2006resilient}
Josh Bongard, Victor Zykov, and Hod Lipson.
\newblock Resilient machines through continuous self-modeling.
\newblock \emph{Science}, 314\penalty0 (5802):\penalty0 1118--1121, 2006.
\newblock URL \url{https://doi.org/10.1126/science.1133687}.

\bibitem[Caluwaerts et~al.(2014)Caluwaerts, Despraz, I{\c{s}}{\c{c}}en,
  Sabelhaus, Bruce, Schrauwen, and SunSpiral]{caluwaerts2014design}
Ken Caluwaerts, J{\'e}r{\'e}mie Despraz, At{\i}l I{\c{s}}{\c{c}}en, Andrew~P
  Sabelhaus, Jonathan Bruce, Benjamin Schrauwen, and Vytas SunSpiral.
\newblock Design and control of compliant tensegrity robots through simulation
  and hardware validation.
\newblock \emph{Journal of the royal society interface}, 11\penalty0
  (98):\penalty0 20140520, 2014.
\newblock URL \url{https://doi.org/10.1098/rsif.2014.0520}.

\bibitem[Cellucci et~al.(2017)Cellucci, MacCurdy, Lipson, and
  Risi]{cellucci20171d}
Daniel Cellucci, Robert MacCurdy, Hod Lipson, and Sebastian Risi.
\newblock 1d printing of recyclable robots.
\newblock \emph{IEEE Robotics and Automation Letters}, 2\penalty0 (4):\penalty0
  1964--1971, 2017.
\newblock URL \url{https://doi.org/10.1109/LRA.2017.2716418}.

\bibitem[Cully et~al.(2015)Cully, Clune, Tarapore, and Mouret]{cully2015robots}
Antoine Cully, Jeff Clune, Danesh Tarapore, and Jean-Baptiste Mouret.
\newblock Robots that can adapt like animals.
\newblock \emph{Nature}, 521:\penalty0 503--507, 2015.
\newblock URL \url{https://doi.org/10.1038/nature14422}.

\bibitem[Golemo et~al.(2018)Golemo, Taiga, Courville, and
  Oudeyer]{pmlr-v87-golemo18a}
Florian Golemo, Adrien~Ali Taiga, Aaron Courville, and Pierre-Yves Oudeyer.
\newblock Sim-to-real transfer with neural-augmented robot simulation.
\newblock In \emph{Proceedings of The 2nd Conference on Robot Learning},
  volume~87 of \emph{Proceedings of Machine Learning Research}, pages 817--828.
  PMLR, 2018.
\newblock URL \url{http://proceedings.mlr.press/v87/golemo18a.html}.

\bibitem[Harvey et~al.(1997)Harvey, Husbands, Cliff, Thompson, and
  Jakobi]{HARVEY1997205}
I.~Harvey, P.~Husbands, D.~Cliff, A.~Thompson, and N.~Jakobi.
\newblock Evolutionary robotics: the sussex approach.
\newblock \emph{Robotics and Autonomous Systems}, 20\penalty0 (2):\penalty0 205
  -- 224, 1997.
\newblock ISSN 0921-8890.
\newblock URL \url{https://doi.org/10.1016/S0921-8890(96)00067-X}.

\bibitem[Hiller and Lipson(2011)]{hiller2011automatic}
Jonathan Hiller and Hod Lipson.
\newblock Automatic design and manufacture of soft robots.
\newblock \emph{IEEE Transactions on Robotics}, 28\penalty0 (2):\penalty0
  457--466, 2011.
\newblock URL \url{https://doi.org/10.1109/TRO.2011.2172702}.

\bibitem[Hiller and Lipson(2014)]{hiller2014dynamic}
Jonathan Hiller and Hod Lipson.
\newblock Dynamic simulation of soft multimaterial 3{D}-printed objects.
\newblock \emph{Soft Robotics}, 1\penalty0 (1):\penalty0 88--101, 2014.
\newblock URL \url{https://doi.org/10.1089/soro.2013.0010}.

\bibitem[Holden et~al.(2019)Holden, Duong, Datta, and
  Nowrouzezahrai]{holden2019subspace}
Daniel Holden, Bang~Chi Duong, Sayantan Datta, and Derek Nowrouzezahrai.
\newblock Subspace neural physics: fast data-driven interactive simulation.
\newblock In \emph{Proceedings of the 18th annual ACM SIGGRAPH/Eurographics
  Symposium on Computer Animation}, page~6. ACM, 2019.
\newblock URL \url{https://doi.org/10.1145/3309486.3340245}.

\bibitem[Hwangbo et~al.(2019)Hwangbo, Lee, Dosovitskiy, Bellicoso, Tsounis,
  Koltun, and Hutter]{hwangbo2019learning}
Jemin Hwangbo, Joonho Lee, Alexey Dosovitskiy, Dario Bellicoso, Vassilios
  Tsounis, Vladlen Koltun, and Marco Hutter.
\newblock Learning agile and dynamic motor skills for legged robots.
\newblock \emph{Science Robotics}, 4\penalty0 (26), 2019.
\newblock URL \url{https://doi.org/10.1126/scirobotics.aau5872}.

\bibitem[Jakobi et~al.(1995)Jakobi, Husbands, and Harvey]{jakobi1995noise}
Nick Jakobi, Phil Husbands, and Inman Harvey.
\newblock Noise and the reality gap: The use of simulation in evolutionary
  robotics.
\newblock In \emph{European Conference on Artificial Life}, pages 704--720.
  Springer, 1995.
\newblock URL \url{https://doi.org/10.1007/3-540-59496-5_337}.

\bibitem[Koos et~al.(2012)Koos, Mouret, and Doncieux]{koos2012transferability}
Sylvain Koos, Jean-Baptiste Mouret, and St{\'e}phane Doncieux.
\newblock The transferability approach: Crossing the reality gap in
  evolutionary robotics.
\newblock \emph{IEEE Transactions on Evolutionary Computation}, 17\penalty0
  (1):\penalty0 122--145, 2012.
\newblock URL \url{https://doi.org/10.1109/TEVC.2012.2185849}.

\bibitem[Kriegman et~al.(2019)Kriegman, Walker, Shah, Levin, Kramer-Bottiglio,
  and Bongard]{kriegman2019automated}
Sam Kriegman, Stephanie Walker, Dylan Shah, Michael Levin, Rebecca
  Kramer-Bottiglio, and Josh Bongard.
\newblock Automated shapeshifting for function recovery in damaged robots.
\newblock In \emph{Proceedings of Robotics: Science and Systems}, 2019.
\newblock URL \url{http://www.roboticsproceedings.org/rss15/p28.html}.

\bibitem[Kwiatkowski and Lipson(2019)]{kwiatkowski2019task}
Robert Kwiatkowski and Hod Lipson.
\newblock Task-agnostic self-modeling machines.
\newblock \emph{Science Robotics}, 4\penalty0 (26), 2019.
\newblock URL \url{http://doi.org/10.1126/scirobotics.aau9354}.

\bibitem[Levin(2013)]{levin2013reprogramming}
Michael Levin.
\newblock Reprogramming cells and tissue patterning via bioelectrical pathways:
  molecular mechanisms and biomedical opportunities.
\newblock \emph{Wiley Interdisciplinary Reviews: Systems Biology and Medicine},
  5\penalty0 (6):\penalty0 657--676, 2013.
\newblock URL \url{https://doi.org/10.1002/wsbm.1236}.

\bibitem[Lipson and Pollack(2000)]{lipson2000automatic}
Hod Lipson and Jordan~B Pollack.
\newblock Automatic design and manufacture of robotic lifeforms.
\newblock \emph{Nature}, 406\penalty0 (6799):\penalty0 974, 2000.
\newblock URL \url{https://doi.org/10.1038/35023115}.

\bibitem[Macklin et~al.(2019)Macklin, Erleben, M{\"u}ller, Chentanez, Jeschke,
  and Makoviychuk]{macklin2019non}
Miles Macklin, Kenny Erleben, Matthias M{\"u}ller, Nuttapong Chentanez, Stefan
  Jeschke, and Viktor Makoviychuk.
\newblock Non-smooth newton methods for deformable multi-body dynamics.
\newblock \emph{ACM Transactions on Graphics (TOG)}, 38\penalty0 (5):\penalty0
  140, 2019.
\newblock URL \url{https://doi.org/10.1145/3338695}.

\bibitem[Majidi et~al.(2013)Majidi, Shepherd, Kramer, Whitesides, and
  Wood]{majidi2013influence}
Carmel Majidi, Robert~F Shepherd, Rebecca~K Kramer, George~M Whitesides, and
  Robert~J Wood.
\newblock Influence of surface traction on soft robot undulation.
\newblock \emph{The International Journal of Robotics Research}, 32\penalty0
  (13):\penalty0 1577--1584, 2013.
\newblock URL \url{https://doi.org/10.1177/0278364913498432}.

\bibitem[Matas et~al.(2018)Matas, James, and Davison]{matas2018sim}
Jan Matas, Stephen James, and Andrew~J Davison.
\newblock Sim-to-real reinforcement learning for deformable object
  manipulation.
\newblock In \emph{Proceedings of The 2nd Conference on Robot Learning},
  volume~87 of \emph{Proceedings of Machine Learning Research}, pages 734--743.
  PMLR, 2018.
\newblock URL \url{http://proceedings.mlr.press/v87/matas18a.html}.

\bibitem[Miglino et~al.(1994)Miglino, Nafasi, and Taylor]{Miglino1994Selection}
Orazio Miglino, Kourosh Nafasi, and Charles~E. Taylor.
\newblock Selection for wandering behavior in a small robot.
\newblock \emph{Artificial Life}, 2\penalty0 (1):\penalty0 101--116, 1994.
\newblock URL \url{https://doi.org/10.1162/artl.1994.2.1.101}.

\bibitem[Moeckel et~al.(2013)Moeckel, Perov, Nguyen, Vespignani, Bonardi,
  Pouya, Sproewitz, van~den Kieboom, Wilhelm, and Ijspeert]{moeckel2013gait}
Rico Moeckel, Yura~N Perov, Anh~The Nguyen, Massimo Vespignani, St{\'e}phane
  Bonardi, Soha Pouya, Alexander Sproewitz, Jesse van~den Kieboom,
  Fr{\'e}d{\'e}ric Wilhelm, and Auke~Jan Ijspeert.
\newblock Gait optimization for roombots modular robots—matching simulation
  and reality.
\newblock In \emph{The IEEE/RSJ International Conference on Intelligent Robots
  and Systems (IROS)}, pages 3265--3272, 2013.
\newblock URL \url{https://doi.org/10.1109/IROS.2013.6696820}.

\bibitem[Nachum et~al.(2019)Nachum, Ahn, Ponte, Gu, and Kumar]{nachum2019multi}
Ofir Nachum, Michael Ahn, Hugo Ponte, Shixiang Gu, and Vikash Kumar.
\newblock Multi-agent manipulation via locomotion using hierarchical sim2real.
\newblock In \emph{Proceedings of The 3rd Conference on Robot Learning}, 2019.
\newblock URL \url{https://arxiv.org/abs/1908.05224}.

\bibitem[Peng et~al.(2018)Peng, Andrychowicz, Zaremba, and Abbeel]{peng2018sim}
Xue~Bin Peng, Marcin Andrychowicz, Wojciech Zaremba, and Pieter Abbeel.
\newblock Sim-to-real transfer of robotic control with dynamics randomization.
\newblock In \emph{The IEEE International Conference on Robotics and Automation
  (ICRA)}, 2018.
\newblock URL \url{https://doi.org/10.1109/ICRA.2018.8460528}.

\bibitem[Pinto et~al.(2018)Pinto, Andrychowicz, Welinder, Zaremba, and
  Abbeel]{Pinto-RSS-18}
Lerrel Pinto, Marcin Andrychowicz, Peter Welinder, Wojciech Zaremba, and Pieter
  Abbeel.
\newblock Asymmetric actor critic for image-based robot learning.
\newblock In \emph{Proceedings of Robotics: Science and Systems}, 2018.
\newblock URL \url{https://doi.org/10.15607/RSS.2018.XIV.008}.

\bibitem[Rosser et~al.(2019)Rosser, Kok, Chahl, and
  Bongard]{rosser2019sim2real}
Kent Rosser, Jia Kok, Javaan Chahl, and Josh Bongard.
\newblock Sim2real transfer degrades non-monotonically with morphological
  complexity for flapping wing design.
\newblock \emph{arXiv preprint}, 2019.
\newblock URL \url{https://arxiv.org/abs/1910.13790}.

\bibitem[Rusu et~al.(2017)Rusu, Večerík, Rothörl, Heess, Pascanu, and
  Hadsell]{rusu2017sim}
Andrei~A. Rusu, Matej Večerík, Thomas Rothörl, Nicolas Heess, Razvan
  Pascanu, and Raia Hadsell.
\newblock Sim-to-real robot learning from pixels with progressive nets.
\newblock In \emph{Proceedings of the 1st Annual Conference on Robot Learning},
  volume~78 of \emph{Proceedings of Machine Learning Research}, pages 262--270.
  PMLR, 2017.
\newblock URL \url{http://proceedings.mlr.press/v78/rusu17a.html}.

\bibitem[Tan et~al.(2018)Tan, Zhang, Coumans, Iscen, Bai, Hafner, Bohez, and
  Vanhoucke]{tan2018sim}
Jie Tan, Tingnan Zhang, Erwin Coumans, Atil Iscen, Yunfei Bai, Danijar Hafner,
  Steven Bohez, and Vincent Vanhoucke.
\newblock Sim-to-real: Learning agile locomotion for quadruped robots.
\newblock In \emph{Proceedings of Robotics: Science and Systems}, 2018.
\newblock URL \url{https://doi.org/10.15607/RSS.2018.XIV.010}.

\bibitem[Tobin et~al.(2017)Tobin, Fong, Ray, Schneider, Zaremba, and
  Abbeel]{tobin2017domain}
Josh Tobin, Rachel Fong, Alex Ray, Jonas Schneider, Wojciech Zaremba, and
  Pieter Abbeel.
\newblock Domain randomization for transferring deep neural networks from
  simulation to the real world.
\newblock In \emph{The IEEE/RSJ International Conference on Intelligent Robots
  and Systems (IROS)}, 2017.
\newblock URL \url{https://doi.org/10.1109/IROS.2017.8202133}.

\bibitem[Zykov et~al.(2007)Zykov, Chan, and Lipson]{zykov2007molecubes}
Victor Zykov, Andrew Chan, and Hod Lipson.
\newblock Molecubes: An open-source modular robotics kit.
\newblock In \emph{IROS Self-Reconfigurable Robotics Workshop}, 2007.
\newblock URL \url{http://molecubes.org}.

\end{thebibliography}

\end{document}